\documentclass[10pt,twocolumn,letterpaper]{article}

\usepackage[pagenumbers]{iccv} 

\definecolor{iccvblue}{rgb}{0.21,0.49,0.74}
\usepackage[pagebackref,breaklinks,colorlinks,allcolors=iccvblue]{hyperref}
\usepackage{multirow}
\usepackage{float} 
\def\confName{ICCV}
\def\confYear{2025}

\usepackage[capitalize]{cleveref}
\crefname{section}{Sec.}{Secs.}
\crefname{table}{Tab.}{Tabs.}
\crefname{figure}{Fig.}{Figs.}

\title{Distilled-3DGS: Distilled 3D Gaussian Splatting}
\author{
Lintao Xiang$^{1,2^*}$ \quad
Xinkai Chen$^{2^*}$ \quad
Jianhuang Lai$^{3}$ \quad
Guangcong Wang$^{2^\dagger}$ \quad 
\vspace{0.2em} \\
$^1$The University of Manchester, \\
$^2$Vision, Graphics, and X Group, Great Bay University, \quad
$^3$Sun Yat-Sen University
}

\begin{document}

\twocolumn[{
\renewcommand\twocolumn[1][]{#1}
\maketitle
\begin{center}
  \vspace{-15pt}
  \centering
  \includegraphics[width=\linewidth]{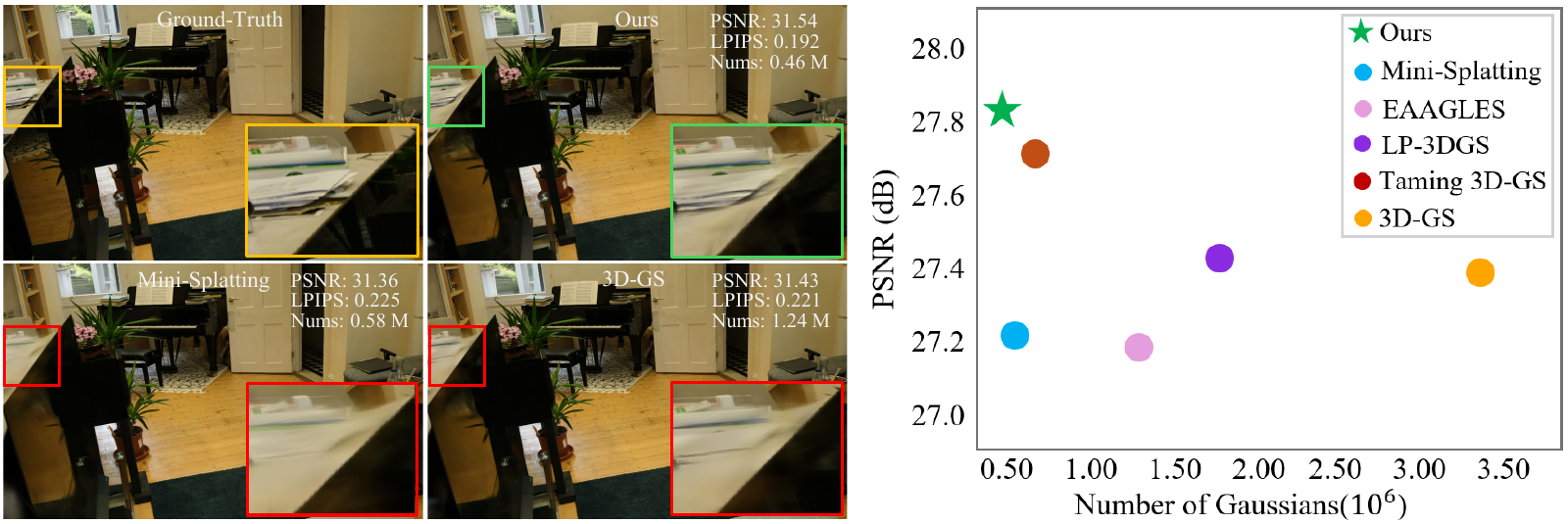}
    \captionof{figure}{Compared with state-of-the-art 3DGS-based methods on Mip360 dataset, our \textbf{Distilled-3DGS} introduces a novel lightweight framework for high-quality view synthesis, achieving better detail preservation with lower storage.}
  \label{fig:teaser} 
\end{center}
}
]

\renewcommand\thefootnote{} %
\footnotetext{$^*$Equal contribution. $^\dagger$Corresponding author.}
\renewcommand\thefootnote{\arabic{footnote}} %

\begin{abstract}
3D Gaussian Splatting (3DGS) has exhibited remarkable efficacy in novel view synthesis (NVS). However, it suffers from a significant drawback: achieving high-fidelity rendering typically necessitates a large number of 3D Gaussians, resulting in substantial memory consumption and storage requirements. To address this challenge, we propose the first knowledge distillation framework for 3DGS, featuring various teacher models, including vanilla 3DGS, noise-augmented variants, and dropout-regularized versions.  The outputs of these teachers are aggregated to guide the optimization of a lightweight student model. To distill the hidden geometric structure, we propose a structural similarity loss to boost the consistency of spatial geometric distributions between the student and teacher model. Through comprehensive quantitative and qualitative evaluations across diverse datasets, the proposed Distilled-3DGS—a simple yet effective framework without bells and whistles—achieves promising rendering results in both rendering quality and storage efficiency compared to state-of-the-art methods. Project page: \href{https://distilled3dgs.github.io/}{https://distilled3dgs.github.io}. Code: \href{https://github.com/lt-xiang/Distilled-3DGS}{https://github.com/lt-xiang/Distilled-3DGS}.

\end{abstract}

\section{Introduction}
\label{sec:intro}
Novel view synthesis (NVS) is a fundamental task~\cite{zhou2016view} in computer vision and computer graphics, serving as a cornerstone in many applications, e.g., VR/AR and autonomous driving. The goal of NVS is to generate photorealistic images from novel, previously unseen viewpoints. This process typically begins by constructing a 3D representation from a set of existing 2D observations. 3D Gaussian Splatting (3DGS)~\cite{kerbl20233d} has recently demonstrated remarkable effectiveness in novel view synthesis. This approach employs a point-based representation augmented with 3D Gaussian attributes and utilizes a rasterization-based rendering pipeline to synthesize images. However, 3DGS necessitates a large number of 3D Gaussians to ensure high-fidelity image rendering, particularly in the presence of complex scenes. This limits their applicability on platforms and devices with constrained computational resources and limited memory capacity.

On the other hand, knowledge distillation~\cite{buciluǎ2006model} has proven effective in compressing neural networks across various vision tasks. However, applying it to 3D Gaussian Splatting (3DGS) introduces unique challenges. First, 3DGS is an explicit and unstructured representation composed of variable 3D Gaussians, lacking the consistent latent feature spaces typically leveraged in conventional KD. Second, the Gaussian primitives are scene-dependent and unordered, preventing straightforward correspondence between teacher and student components. Third, since rendering outputs are view-dependent and non-differentiable w.r.t. individual Gaussians, designing stable and informative distillation losses becomes nontrivial. These challenges necessitate careful design of both teacher model ensembles and geometry-aware distillation strategies, as introduced in our Distilled-3DGS framework.

To address this challenge, we propose a lightweight 3D Gaussian representation framework based on knowledge distillation, termed Distilled-3DGS. This approach enhances the performance of a compact student model by distilling knowledge from multiple complex teacher models. The overall pipeline of Distilled-3DGS comprises two main stages: multi-teacher training and student training via distillation. In the multi-teacher training stage, we begin by training a standard 3DGS model. Subsequently, we introduce random perturbations and dropout strategies separately to obtain two additional diverse teacher models. During the distillation-based student training stage, we first aggregate predictions from the teacher ensemble to synthesize a pseudo image. The student model is then supervised by enforcing similarity between its rendered output and this pseudo image. This strategy effectively transfers rich knowledge priors from the teacher models, providing a more comprehensive and robust supervisory signal for optimizing the student model.

In the context of Distilled-3DGS, the teacher model typically contains dense and high-quality point clouds, while the student model is compressed to obtain much sparser points due to efficiency or deployment considerations. Despite its sparsity, the student model is trained to reconstruct the same underlying 3D scene as the teacher. So it is expected to preserve the essential spatial layout and local geometric patterns present in the teacher model. To facilitate this, we propose a spatial distribution distillation strategy that guides the student to align with the teacher’s point distribution in space. Rather than enforcing exact point-wise correspondence, this structure-aware supervision encourages the student to learn how the teacher organizes points, focusing on global and local geometric consistency. In this way, structural knowledge from the teacher can be effectively and comprehensively distilled into the student model.

In summary, our main contributions are as follows: 1) We propose a novel distillation-based 3DGS framework, termed as Distilled-3DGS, which is the first method to leverage multi-teacher knowledge priors to optimize 3DGS and boost rendering quality and storage efficiency. 2) We propose a spatial distribution consistency distillation to enable the student model to learn similar geometric structure distributions from the teacher model. 3) Extensive experiments on several real-world datasets—including Mip-NeRF 360, Tanks\&Temples, and Deep Blending—demonstrate that the proposed Distilled-3DGS achieves promising performance in both rendering quality and efficiency compared to existing methods.
\section{Related Work}
\label{sec:related work}

\noindent\textbf{3D Representation.} Radiance fields have been extensively employed for 3D scene reconstruction, particularly in the context of novel view synthesis. Neural Radiance Fields (NeRFs) have achieved remarkable progress by learning neural volumetric representations of 3D scenes, enabling high-fidelity image synthesis via volumetric rendering techniques. After that, many works have focused on improving the rendering quality~\cite{barron2021mip,barron2022mip} and accelerating the efficiency~\cite{fridovich2022plenoxels,hu2023tri,zhao2023instant,yu2021plenoctrees} of NeRFs. Nevertheless, NeRF-based approaches continue to rely on numerous MLP queries during rendering, thereby limiting their applicability in scenarios with real-time constraints. To enhance the training and rendering efficiency, Plenoxels~\cite{fridovich2022plenoxels} improve NeRF efficiency by optimizing a sparse voxel grid and removing the need for MLPs, while Instant NGP~\cite{zhao2023instant} uses hash-grid encodings to boost expressivity. However, despite these improvements, grid-based methods still struggle to achieve real-time rendering. Recently, 3D Gaussian Splatting (3DGS)~\cite{kerbl20233d} has gained significant attention as an efficient and effective approach for 3D scene representation. 3DGS represents 3D scenes explicitly with millions of anisotropic Gaussians and utilizes differentiable rasterization, enabling real-time, photorealistic view synthesis. When 3DGS overfits the scene by optimizing Gaussian properties, it typically produces many redundant Gaussians, thereby reducing rendering efficiency and substantially increasing memory usage. 

To tackle these issues, several subsequent approaches have aimed to prune redundant Gaussians based on hand-crafted importance criteria. Mini-Splatting~\cite{fang2024mini} addresses overlapping and reconstruction artifacts by employing blur splitting, depth reinitialization, and stochastic sampling. Radsplatting~\cite{niemeyer2024radsplat} enhances robustness by applying a max operator to derive importance scores from ray contributions. Taming-3DGS~\cite{mallick2024taming} leverages pixel saliency and gradient information for selective densification, while LP-3DGS~\cite{zhang2024lp} utilizes a learned binary mask for efficient Gaussian pruning. Additionally, Scaffold-GS~\cite{lu2024scaffold} proposes a structured dual-layer hierarchical scene representation to better regulate the distribution of 3D Gaussian primitives. 

Overall, the aforementioned methods that prioritize efficiency generally achieve faster performance, but this often comes at the expense of rendering quality compared to the standard 3DGS. Conversely, approaches that focus on enhancing rendering quality tend to demand substantially higher computational resources. To address this trade-off, we propose a knowledge distillation-based 3DGS framework that simultaneously improves storage efficiency and rendering fidelity.

\noindent\textbf{Knowledge Distillation.}Knowledge distillation (KD) transfers knowledge from a large teacher model to a compact student model. Initially proposed for model compression~\cite{buciluǎ2006model,hinton2015distilling}, KD began with matching teacher outputs and was later extended to mimic intermediate representations~\cite{zagoruyko2016paying,tian2019contrastive}. KD has since been applied to various tasks, including detection~\cite{li2017mimicking,zhang2020improve},segmentation~\cite{liu2019structured,ji2022structural}, and generation~\cite{li2020gan,zhang2022wavelet}. To overcome the limitations of single-teacher KD, multi-teacher distillation (MKD) is proposed to aggregate diverse knowledge from multiple teachers. While early approaches assign equal weights~\cite{you2017learning,fukuda2017efficient}, recent methods adopt adaptive strategies, such as entropy-based weighting (EB-KD~\cite{kwon2020adaptive}) and confidence-aware distillation (CA-MKD~\cite{zhang2022confidence}). MMKD~\cite{zhang2023adaptive} further introduces meta-learning to jointly distill features and logits. These methods often rely on CNNs for structured feature spaces, facilitating effective alignment via soft labels or intermediate supervision.

However, extending KD to 3D Gaussian Splatting (3DGS) poses new challenges. 3DGS uses an explicit and unstructured representation composed of a variable set of discrete Gaussian primitives. These primitives are unordered, scene-dependent, and lack a shared latent space, making it infeasible to directly align elements between teacher and student. As a result, existing KD strategies must be fundamentally rethought to accommodate the unique properties of 3DGS. Based on the above analysis, we propose to utilize multiple pre-trained 3DGS teacher models to render high-quality images as supervision targets and optimize the student model. Besides, we propose a spatial distribution distillation strategy that guides the student to align with the teacher’s point distribution in space.

\section{Method}
\label{sec:method}

In this section, we present Distilled-3DGS, an efficient 3D Gaussian Splatting framework that distills knowledge from powerful teacher models to a small student model. 

\subsection{Preliminaries}
3DGS~\cite{kerbl20233d} is a cutting-edge method for novel view synthesis, which fundamentally depends on an explicit point-based representation to achieve high-fidelity rendering from arbitrary viewpoints. Specifically, 3DGS models a scene as a set of Gaussian distributions. The $i_{th}$ Gaussian primitive is denoted as $G_i=(\mu_i, \Sigma_i, o_i, f_i)$, where $\mu_i$ is the 3D position, $\Sigma_i$ is the covariance matrix, $f_i$ represents spherical harmonics (SH) coefficients associated with the Gaussian, and $o_i$ indicates opacity. The effect of the $i_{th}$ Gaussian primitive at position $x$ is represented by $G_i(x) = e^{-\frac{1}{2}(x-\mu_i)^T \Sigma^{-1}_i(x-\mu_i)}$, where $\Sigma_i$ can be factorized as $\Sigma_i = RSS^TR^T$, with $R$ as a rotation matrix and $S$ as a scaling matrix, both being learnable. Subsequently, the Gaussians are mapped to the 2D image plane via the projection matrix $W$, resulting in the projected 2D covariance matrix ${\Sigma^{'}_i  = JW \Sigma_i W^{T}J^{T}}$, where $J$ represents the Jacobian of the affine projection. The pixel color is computed through alpha blending as follows:

\begin{equation}
\mathbf{c} = \sum_{i=1}^{N} c_i \alpha_i \prod_{j=1}^{i-1} (1 - \alpha_j),
\end{equation}
where $N$ is the number of Gaussians covering the pixel, the color $c_i$ is derived from the spherical harmonics (SH) coefficients of each Gaussian, while $\alpha_i$ is determined by the projected 2D covariance matrices $\Sigma^{'}$ and the associated opacity $o_i$. The Gaussian parameters are optimized using a photometric loss~\cite{kerbl20233d} function, with the posed training images providing the ground-truth supervision.

\subsection{Distilled 3D Gaussian Splatting}
3DGS has enabled highly detailed and accurate 3D scene reconstruction, yet such state-of-the-art models are often extremely large and computationally expensive, limiting their practicality in real-time and resource-constrained scenarios. 
Knowledge distillation (KD) has emerged as a highly effective and popular approach for model compression in image classification, semantic segmentation~\cite{liu2019structured}, and object detection~\cite{chen2017learning}. Inspired by these observations, one could ask if knowledge distillation works for 3DGS. Different from conventional KD in neural networks, it is an explicit 3D representation with variable unstructured 3D Gaussians, which remains unexplored. To address this problem, we first provide an overview of the proposed Distilled-3DGS, and then detail the design of diverse teacher models and the distillation method, as discussed in the following.

\begin{figure*}[htbp]
\begin{center}
    \includegraphics[width=0.9\textwidth]{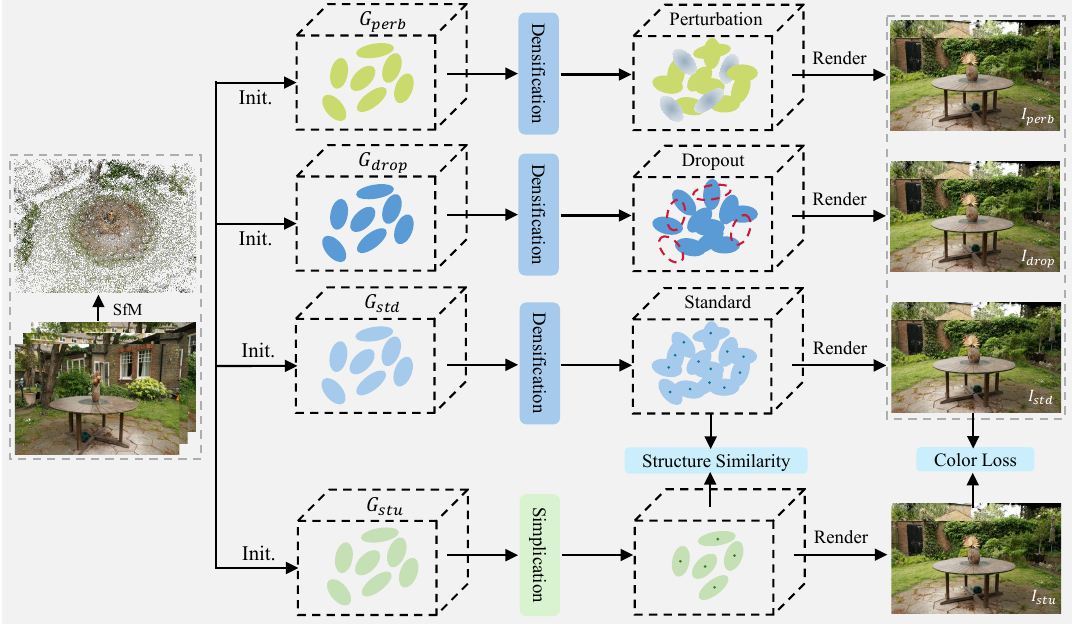} \\
    \caption{The architecture of multi-teacher knowledge distillation framework for 3DGS. It consists of two stages. First, a standard teacher model $G_{std}$ is trained, along with two variants: $G_{perb}$ with random perturbation and $G_{drop}$ with random dropout. Then, a pruned student model $G_{std}$ is supervised by the outputs of these teachers. Additionally, a spatial distribution distillation strategy is introduced to help the student learn structural patterns from the teachers.}
    \label{fig:pipeline}
\end{center}
\end{figure*}

\subsection{Overview of Distill-3DGS}
As shown in Fig.\ref{fig:pipeline}, we firstly train three independent 3DGS models with diverse strategies to obtain cumbersome yet high-quality teacher models with millions of Gaussian primitives, optimized by the standard photometric loss. Then we leverage the optimized teacher 3DGS representation to generate pseudo labels by fusing multiple teachers' outputs. In the training process of the student model, pseudo labels are leveraged to transfer prior knowledge from multiple teachers to a single student. To obtain a lightweight student model, we prune the number of Gaussians based on the importance score proposed in Mini-Splatting~\cite{fang2024mini}. To distill knowledge hidden in geometric structure of 3D Gaussians, we propose structural knowledge distillation for unstructured 3D Gaussians to encourage the similar spatial distribution between teacher and student models. 

\subsection{Training Diverse Teacher models } 
To provide the student model with richer supervision signals and facilitate a better understanding of 3D scene structures and details, we train the base 3DGS model multiple times using diverse strategies to enhance the robustness and generalization ability of the teacher models.

\noindent\textbf{Regular training.} First, we train a vanilla 3DGS model $G_{std}$ with the same settings in~\cite{kerbl20233d}. The training loss is defined as: 
\begin{equation}
\mathcal{L}_{\text{color}} = (1 - \lambda) \mathcal{L}_1 (\hat{I}, I_{gt}) + \lambda \mathcal{L}_{\mathrm{D\text{-}SSIM}} (\hat{I}, I_{gt})
\end{equation}

\noindent\textbf{Feature Perturbation.} Then, we train a 3DGS model $G_{perb}$ with random perturbations on Gaussian parameters, following the same optimization and density control strategies introduced in 3DGS. At each training iteration $t$, each Gaussian is perturbed as:
\begin{equation}
    G^t_{perb} = G^t_{std} + \delta_t,
\end{equation}
where random noises are added to corresponding Gaussian parameters including 3D positions $\mu_p$, 3D rotations $R_p$, scales $S_p$, and opacities $o_p$. Since the representation of rotation as a 3D matrix is discontinuous, we instead perturb its continuous 6D representation as :

\begin{equation}
    \hat{\mathbf{R}}_{p}^{t} = f^{-1}\left( f(\mathbf{R}_{p}^{t}) + \delta_t^R\, \right), \quad \delta_p^t \in \mathbb{R}^6
\end{equation}
where $f$ and $f^{-1}$ are the forward and inverse mappings between the rotation matrix and its 6D representation. By introducing parameter perturbations during training, the model is compelled to learn scene structures that are less dependent on the precise positions and shapes of Gaussian primitives, thereby enhancing its generalization capability.

\noindent\textbf{Random Dropout.} Dropout~\cite{gal2016dropout, srivastava2014dropout} is recognized as one of the most effective techniques for improving model robustness by randomly deactivating a subset of neurons during training. Inspired by the remarkable success of dropout, we introduce a Random Dropout Strategy to further enhance both the robustness and redundancy of the representation of  model $G_{drop}$. Specifically, during training, each Gaussian primitive is randomly deactivated with probability $p$, while the remaining primitives are optimized to fit the observed views. During inference, all Gaussian primitives are activated to facilitate novel view synthesis. By randomly dropping a subset of Gaussian primitives during training, our approach encourages the model to learn a collaborative and distributed scene representation, rather than relying on a limited set of critical Gaussian primitives. Inspired by ~\cite{park2025dropgaussian}, the dropping rate $r_{t}$ is updated based on the current iteration index $t$ as follows:

\begin{equation}
    r_{t} = r_{init} \cdot (t - t_0) / (t_1 - t_0),
\end{equation}
where $t_0$ and $t_1$ are the starting and end iterations of introducing random dropout strategy. $r_{init}$ is the initial drop rate.

\subsection{Training Efficient Student Model}
Knowledge distillation is a technique that transfers knowledge from a larger teacher model to a smaller, faster student model. This approach is particularly useful when deploying deep neural networks in resource-constrained environments. The student model, trained under the guidance of the teacher, can achieve comparable performance with significantly fewer parameters. This process mainly consists of pseudo label generation and student training.

\noindent\textbf{Pseudo labeling with teacher model.} By evaluating the optimized diverse teachers $G_{std}$, $G_{perb}$ and $G_{drop}$, we can render per-view image denoted as $I_{std}$, $I_{perb}$ and $I_{drop}$, these rendered images as prior knowledge are further aggregated by average strategy to generate pseudo label $I_{tea}$ and guide the learning of the student model.

\noindent\textbf{Conventional Knowledge Distillation.}  In the training process of the student network, we utilize the ground-truth labels and the pseudo label of multiple teachers as additional knowledge to jointly guide the optimization of student model $G_{stu}$. Following the conventional knowledge distillation loss, we formulate our objective by incorporating fused knowledge from multiple teachers, as follows:
\begin{equation}
\mathcal{L}_{\text{kd}} = \mathcal{L}_{color} (I_{stu}, I_{gt}) + \lambda_{kd} \mathcal{L}_{color} (I_{stu}, I_{tea})
\end{equation}

\noindent\textbf{Spatial Distribution Distillation.} In the context of 3DGS, these optimized teacher models provide a structure-rich and dense 3D point distribution. In contrast, the student model operates under sparse or limited sampling conditions and aims to reconstruct the similar scene representation. Therefore, we hope to design a structural similarity loss to encourage the student model to capture spatial geometric distributions similar to those of the teacher. However, challenges arise due to varying point densities, sampling noise, and non-uniform point distributions between student and teacher models. Direct coordinate-based distance measures are often insufficiently robust to these variations. 

\begin{figure}[htbp]
\begin{center}
\includegraphics[width=0.45\textwidth]{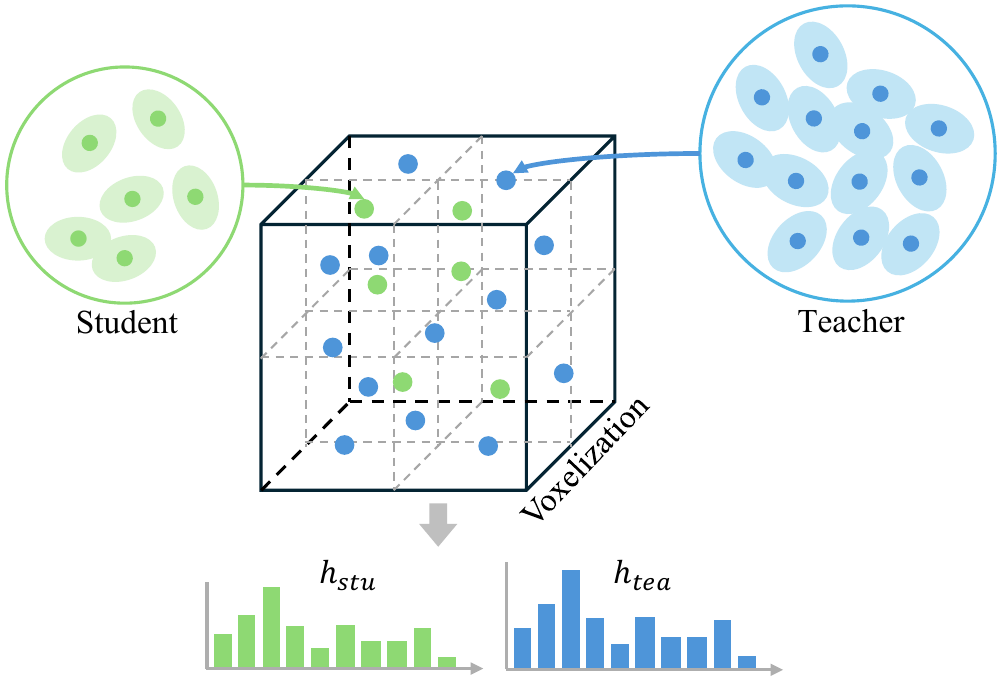}\\
    \caption{Overview of Spatial Distribution Distillation. } 
    \label{fig:voxel}
\end{center}
\end{figure}

To address this problem, we leverage the voxel histogram representation shown in Fig.~\ref{fig:voxel}, which discretizes the 3D space into regular voxels and counts the number of points within each voxel. This approach encodes the spatial distribution of points as a high-dimensional structural feature, inherently robust to point count and density variations. Comparing voxel histograms thus enables efficient and structure-aware similarity evaluation between different point clouds. To this end, we propose a voxel histogram-based structural loss to enhance the structural learning capability of the student model.

During the training phase of the student model, we firstly obtain point cloud $P_{tea}$ and $P_{stu}$ from the optimized standard teacher model $G_{std}$ and student model $G_{stu}$. Then, we determine a common 3D bounding box that encompasses both sets of points.The bounding box is partitioned into a regular voxel grid with a resolution of 128 . Each point from both clouds is assigned to a corresponding voxel cell based on its spatial coordinates. For point cloud $P_{tea}$ and $P_{stu}$, we count the number of points falling into every voxel separately, resulting in two high-dimensional voxel occupancy histograms $\mathbf{h}_{tea}$ and $\mathbf{h}_{stu}$. These histograms are then normalized to form probability distributions that capture the spatial structure of each cloud, independent of point count or density. Finally, we compute the cosine similarity between their normalized voxel occupancy histograms. The cosine similarity loss is given by:
\begin{equation}
\mathcal{L}_{\text{hist}} = 1 - \frac{ \mathbf{h}_{tea} \cdot \mathbf{h}_{stu} }{ \| \mathbf{h}_{tea} \|_2 \, \| \mathbf{h}_{stu} \|_2 }
\end{equation}

This loss quantitatively reflects how closely the Student point cloud matches the Teacher’s structural distribution. The final loss function during the student training phase is defined as:
\begin{equation}
\mathcal{L} = \mathcal{L}_{\text{kd}} + \mathcal{L}_{\text{hist}}
\end{equation}
\section{Experiments}

\noindent\textbf{Datasets.} We conducted experiments on three widely used datasets: LLFF~\cite{mildenhall2019local}, Mip360~\cite{barron2022mip} and two scenes from the Tanks and Temples(T\&T)~\cite{knapitsch2017tanks}. LLFF contains eight scenes with forward-facing camera. Mip-NeRF360 comprises nine distinct scenes that encompass both expansive outdoor scenes and intricate indoor settings. These scenes exhibit a wide range of capture styles and encompass both bounded indoor environments as well as large, unbounded outdoor settings. To partition the dataset into training and test sets, we follow the protocol of 3DGS by allocating every eight image to the test set. The resolution of all images is kept consistent with that used in 3DGS.

\begin{figure*}[htbp]
\begin{center}
\includegraphics[width=1.0\textwidth]{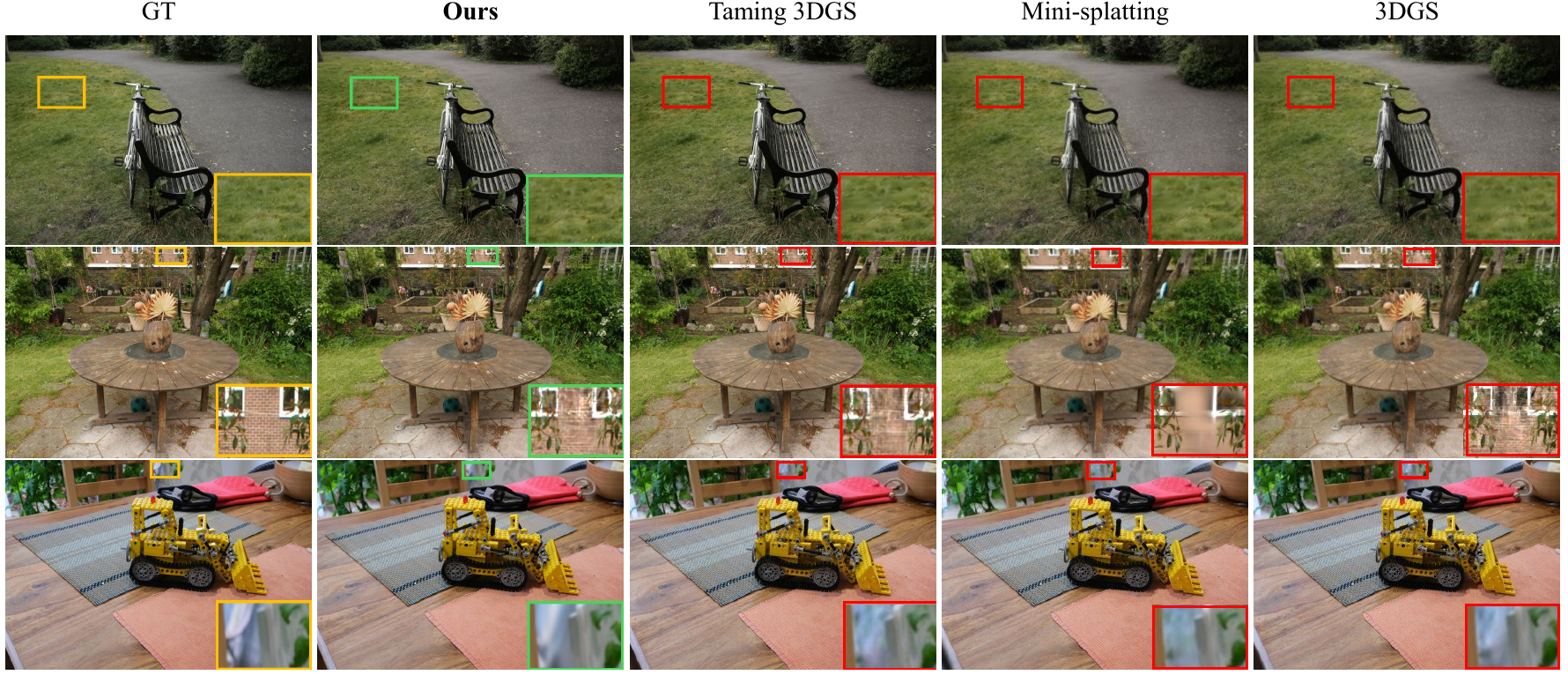}\\
    \caption{Visualized comparison on the \textit{Bicycle}, \textit{Garden}, and \textit{Kitchen} scenes. As shown in the rendered images and corresponding local regions, the proposed method can better capture fine details.
  }
   \label{fig:mipnerfpic}
\end{center}
\end{figure*}

\begin{table*}[htbp]
\centering
\resizebox{2.1\columnwidth}{!}{
\begin{tabular}{l|l|cccc|cccc|cccc}
\hline
\multirow{2}{*}{Type} & \multirow{2}{*}{Method} & \multicolumn{4}{c|}{Mip-NeRF 360} & \multicolumn{4}{c|}{Tanks \& Temples} & \multicolumn{4}{c}{Deep Blending} \\
\cline{3-14}
 & & PSNR$\uparrow$ & SSIM$\uparrow$ & LPIPS$\downarrow$ & \#G($10^6$)$\downarrow$ & PSNR$\uparrow$ & SSIM$\uparrow$ & LPIPS$\downarrow$ & \#G($10^6$)$\downarrow$ & PSNR$\uparrow$ & SSIM$\uparrow$ & LPIPS$\downarrow$ & \#G($10^6$)$\downarrow$ \\
\hline
\multirow{6}{*}{Quality}      
 & Plenoxels (CVPR'22)     & 23.08 & 0.626 & 0.463 & -   &21.08       &0.719       &0.379       &-     &23.06       &0.795       &0.510       &-     \\
    
 & INGP-Big (SIGGRAPH'22)      & 25.59 & 0.699 & 0.331 & -   &21.92       &0.745       &0.305       &-     &24.96       & 0.817       &0.390       &-    \\
   
 & Mip-NeRF360 (CVPR'22)   & 27.69 & 0.792 & 0.237 & -   &22.22       &0.759       & 0.257       &-     &29.40       & 0.901       & 0.245      &-     \\
    
 & 3D-GS (TOG'23)         & 27.26 & 0.815 & 0.214 &3.5     &23.14        &0.841       &0.183       &2.0     &29.41       &0.903       &0.243       &3.2     \\
  & 3D-GS$^*$         & 27.39 & 0.819 & 0.219 &3.43     &23.61        &\underline{0.849}       &0.180       &1.84     &29.55       &\underline{0.912}       &\underline{0.241}       &3.24     \\
 & ScaffoldGS (CVPR'24)    &27.60       &0.812       &0.222       &0.6     &\textbf{24.08}       &\textbf{0.854}       &\textbf{0.165}       &0.6     &\textbf{30.25}       &0.907       &0.245       &0.40     \\

\hline
\multirow{7}{*}{Efficiency}

 & CompactGaussian (CVPR'24) &27.08       & 0.798        &0.247       &1.388     &23.32       &0.831        &0.201       &0.836     &29.79       &0.901       &0.258       &1.06     \\

 & LP-3DGS (NIPS'24) &27.47       & 0.812        &0.227       &1.959     &23.60       &0.842        &0.188       &1.244     &-       &-       &-       &-     \\
 
 & MiniSplatting (CVPR'24) &27.25       & 0.820        &0.217       &\underline{0.5}     &23.21       &0.836        &0.203       &0.32     &\underline{29.98}       &0.908       &0.253       &0.40     \\

 & EAGLES(ECCV'24) &27.20       & 0.809        &0.232       &1.3     &23.26       &0.837        &0.201       &0.7     &29.86       &0.910       &0.246       &1.20     \\

 & Taming 3DGS(SIGGRAPH Asia'24)   &\underline{27.71}       &\underline{0.820}       &\underline{0.207}       &0.63     &\underline{23.95}       &0.837       &0.201       &\underline{0.29}     &29.82       & 0.904       &\textbf{0.237}       &\textbf{0.27}     \\
 

 & CompGS (ECCV'24)   &27.12 &0.806 &0.240 &0.84     
    &23.44 &0.838 &0.198 &0.52
    &29.90 &0.907 &0.251 &0.55     \\

 & \textbf{Ours}       &\textbf{27.81}       &\textbf{0.827}       &\textbf{0.202}       &\textbf{0.49}     &23.76       &0.845       &\underline{0.179}       &\textbf{0.25}  &29.87       &\textbf{0.916}       &0.251       &\underline{0.33}     \\
\hline
\end{tabular}
}
\caption{Quantitative evaluations across the Mip-NeRF 360, Tanks\&Temples, and Deep Blending datasets. \textbf{Best} and \underline{second-best} results are highlighted for each. * denotes our re-runs of the existing codebase to ensure a fair evaluation.}
\label{tab:overall}
\end{table*}

\noindent\textbf{Evaluation Metrics.} For the evaluation of comparative view synthesis quality, we adopt several widely used quantitative metrics, including Peak Signal-to-Noise Ratio (PSNR), Learned Perceptual Image Patch Similarity (LPIPS)\cite{zhang2018unreasonable}, and Structural Similarity Index Measure (SSIM)\cite{wang2004image}. PSNR and SSIM primarily assess pixel-level fidelity and structural consistency, respectively, while LPIPS reflects a more human-aligned assessment of visual quality. In addition, we present the average memory usage associated with storing the optimized parameters.

\noindent\textbf{Implementations.} Our implementations are based on the official 3DGS codebase. All models were trained on a single NVIDIA RTX 3090 GPU. Training process of Distilled-3DGS contains two stages: In the training phase of teacher models, these teacher models are trained for 30k iterations by following 3DGS. Gaussian densification is stopped after the 15000th iteration as in 3DGS. For the teacher training with random perturbation, random noise $\delta_t$ is applied to those Gaussian primitives exhibiting large view-space positional gradients, as these typically correspond to regions that have not yet been well reconstructed, starting from 500th to 15000th iteration with interval 500. Introducing appropriate perturbations in this manner can enhance the robustness of the model. For the teacher training with random dropout, $r_{init}$, $t_0$ and $t_1$ are respectively set 0.2, 500 and 15000. Each Gaussian primitive is randomly deactivated with probability $p$, and the remaining primitives are optimized to fit the observed views. For inference, all Gaussian primitives are activated to enable novel view synthesis.

In the training phase of student model, the total number of optimization steps is set to 30K. Densification is applied up to the 15000th iteration, after which simplification is carried out at both the 15000th iterations. Subsequently, the structural similarity loss is computed and applied every 500 iterations throughout the optimization process.

\subsection{Comparisons with State-of-the-Arts}
We evaluate model performance across several real-world datasets, including Mip-NeRF 360, Tanks\&Temples, and Deep Blending. For NeRF-based methods, we compare with the state-of-the-art Mip-NeRF 360~\cite{barron2022mip} and two efficient NeRF variants, INGP~\cite{muller2022instant} and Plenoxels~\cite{fridovich2022plenoxels}. For 3DGS-based methods, our comparisons include the vanilla 3DGS as well as leading Gaussian simplification techniques such as CompactGaussian~\cite{lee2024compact}, LP-3DGS~\cite{zhang2024lp}, EAGLES~\cite{girish2024eagles}, MiniSplatting~\cite{fang2024mini}, and Taming 3DGS~\cite{mallick2024taming}. For vanilla 3D-GS, we include both the metrics reported in ~\cite{kerbl20233d} and those obtained through our own experimental runs. The quantitative results for all datasets are presented in Table~\ref{tab:overall}. Our method surpasses both the voxel grid-based approach, Plenoxels, and the fast NeRF-based method, INGP, across all datasets and evaluation metrics. Compared to the Mip-NeRF360 baseline, Distilled-3DGS yields PSNR improvements of 0.12 dB, 1.54 dB, and 0.47 dB on the Mip-NeRF360, Tanks\&Temples, and Deep Blending datasets, respectively, verifying its effectiveness across diverse datasets.

Compared to other 3DGS-based methods, our proposed Distilled-3DGS consistently outperforms the baseline 3DGS across all three metrics while using significantly fewer Gaussians on Mip360 dataset. We attribute this improvement to the comprehensive knowledge supervision provided by the diverse teacher models. Specifically, compared to the vanilla 3DGS, our Distilled-3DGS achieves PSNR improvements of 0.55 dB on Mip-NeRF360, 0.62 dB on Tanks\&Temples, and 0.46 dB on Deep Blending. The number of Gaussians is also reduced by 86.0\%, 87.5\%, and 89.6\% on these three datasets, respectively. Compared with these 3DGS simplification methods, such as Taming 3DGS, our method also can improve rendering quality while maintaining a comparable number of Gaussians. Besides, visual comparison is illustrated in Fig.~\ref{fig:mipnerfpic}, compared with existing simplification approaches such as Taming-3DGS, Mini-Splatting, and the vanilla 3DGS, our Distilled-3DGS achieves rendering results that preserve fine details most faithfully to the ground truth while utilizing a significantly reduced number of Gaussian primitives.

\subsection{Ablation Studies and Further Analyses}
To study the contribution of each component in the proposed framework, we conducted a series of ablation experiments on the Deep Blending and Tanks\&Temples datasets.

\noindent\textbf{Effect of the number of teachers.} To further understand the contribution of each teacher model in our distillation framework, we conducted ablation studies by gradually removing the Perturbation-based, Dropout-based 3DGS teachers, respectively. The quantitative results are shown in Table~\ref{tab:abla_tea}. Specifically, the student model distilled from all three teachers consistently achieves the best performance, indicating that each teacher provides complementary knowledge. Teacher $G_{perb}$ enhances the student’s robustness to input variations, while teacher $G_{drop}$ prevents overfitting and encourages generalization. The regular 3DGS teacher $G_{std}$ serves as a strong baseline, providing high-fidelity supervision. The progressive decrease in performance with the removal of these specialized teachers underscores their critical roles in enriching the distilled knowledge, validating the effectiveness of leveraging diverse teacher models for optimal student performance.

\begin{figure}[H]
  \centering
    \resizebox{1.0\columnwidth}{!}{
    \begin{tabular}{l|ccc|ccc}
    \toprule
    \multirow{2}{*}{\textbf{Method}} & \multicolumn{3}{c|}{\textbf{Deep Blending}} & \multicolumn{3}{c}{\textbf{Tanks\&Temples}} \\
     & PSNR$\uparrow$ & SSIM$\uparrow$ & LPIPS$\downarrow$ & PSNR$\uparrow$ & SSIM$\uparrow$ & LPIPS$\downarrow$ \\
    \midrule
    \textbf{Ours} & \textbf{29.87} & \textbf{0.916} & \textbf{0.251} & \textbf{23.76} & \textbf{0.845} & \textbf{0.179} \\
    \midrule
    Without $G_{drop}$ &29.71 &0.899 &0.257 &23.58 &0.840 &0.186 \\
    Without $G_{perb}$ &29.63 &0.878 &0.262 &23.43 &0.838 &0.195 \\
    Without $\mathcal{L}_{\text{hist}}$ &29.47 &0.871 &0.263 &23.32 &0.836 &0.197 \\
    \bottomrule
    \end{tabular}
    }
  \captionof{table}{Ablation study on two datasets.}
  \label{tab:abla_tea}

  \vspace{0.5em}

    \resizebox{0.9\columnwidth}{!}{
    \begin{tabular}{ccccc}
    \toprule
    Grid\_size & PSNR↑ & SSIM↑ & LPIPS↓ & Train Mem.(MB) \\
    \midrule
    32 &27.51 &0.819 &0.198	&9564 \\
    64 &27.62 &0.821 &0.199	&10232  \\
    128 &27.81 &0.827 &0.202 &12235  \\
    256 &27.92 &0.829 &0.203 &15456 \\
    \bottomrule
    \end{tabular}
    }
    \captionof{table}{The impact of different grid size in spatial distribution distillation.}
    \label{tab:grid_size}
    
    \vspace{0.5em}
    
    \resizebox{1.0\columnwidth}{!}{
    \begin{tabular}{l|cccc}
    \toprule
    \multirow{2}{*}{\textbf{Method}} & \multicolumn{4}{c}{\textbf{Room (Mip360)}} \\
    \cmidrule(lr){2-5} 
     & PSNR↑ & SSIM↑ & LPIPS↓ & \#G($10^6$)$\downarrow$  \\
    \midrule
    Teachers ($G_{std}$+$G_{drop}$+$G_{perb}$) &\textbf{32.15} &\textbf{0.935}	&\textbf{0.185} &\textbf{1.56} 	 \\
    \midrule
    3DGS &31.59 &0.920 &0.200 &1.50      \\
    \midrule
    Student(Base) &31.54 &0.927	&0.193 &0.46 \\
    Student(Big) &31.89	&0.934	&0.189  &1.13   \\
    Student(Small) &31.39 &0.923 &0.194  &0.21  \\
    \bottomrule
    \end{tabular}
    }
    \caption{The impact of the number of Gaussians.}
    \label{tab:abla_num}

    \vspace{0.5em}
    
    \resizebox{1.0\columnwidth}{!}{
    \begin{tabular}{l|c|llll}
    \hline
    \multicolumn{1}{c|}{\multirow{2}{*}{\textbf{Teacher}}} & \multicolumn{1}{l|}{\multirow{2}{*}{\textbf{Student}}} & \multicolumn{4}{c}{\textbf{Room (Mip360)}}                                                                \\ \cline{3-6} 
    \multicolumn{1}{c|}{}                         & \multicolumn{1}{l|}{}                         & \multicolumn{1}{c|}{PSNR$\uparrow$ } & \multicolumn{1}{c|}{SSIM$\uparrow$ } & \multicolumn{1}{l|}{LPIPS$\downarrow$} & \#G($10^6$)$\downarrow$ \\ \hline
    $G_{std}/G_{perb}/G_{drop}$                                             & \multirow{5}{*}{$G_{stu}$}                           & \multicolumn{1}{l|}{\textbf{31.54}}     & \multicolumn{1}{l|}{\textbf{0.927}}     & \multicolumn{1}{l|}{\textbf{0.193}}      &0.460     \\ 
    $G_{std}/G_{std}/G_{std}$                                             &                                               & \multicolumn{1}{l|}{31.36}     & \multicolumn{1}{l|}{0.923}     & \multicolumn{1}{l|}{0.191}      &0.469     \\ 
    $G_{std}$                                             &                                               & \multicolumn{1}{l|}{31.19}     & \multicolumn{1}{l|}{0.918}     & \multicolumn{1}{l|}{0.187}      &0.465     \\ 
    $G_{perb}$                                             &                                               & \multicolumn{1}{l|}{31.31}     & \multicolumn{1}{l|}{0.921}     & \multicolumn{1}{l|}{0.186}      &0.453     \\ 
    $G_{drop}$&                                               & \multicolumn{1}{l|}{31.23}     & \multicolumn{1}{l|}{0.919}     & \multicolumn{1}{l|}{0.189}      &0.459   \\ \hline
    \end{tabular}
    }
    \caption{The impact of different teachers}
     \label{tab:abla_combine}
\end{figure}

\noindent\textbf{Effect of Spatial Distribution Distillation.} The results presented in Table~\ref{tab:abla_tea} verify that spatial distribution distillation plays a crucial role in enhancing rendering quality. Without this, performance in \textit{PSNR} is decreased by 0.16 dB. In addition, we investigate the impact of varying grid sizes on the student model training using the Mip360 dataset shown in Table~\ref{tab:grid_size}. Generally, a larger grid size produces smaller voxel dimensions and a greater number of voxels, leading to a more detailed scene representation. While increasing the grid size can improve PSNR performance, it also incurs a substantial increase in GPU memory.

\noindent\textbf{Impact of the number of Gaussians.} We conduct experiments on the \textit{Room} scene from Mip360 to evaluate the impact of Gaussian count. Table~\ref{tab:abla_num} reports the reconstruction quality and the number of Gaussians for different model variants. The ensemble of three diverse teacher models achieves the highest PSNR of 32.15 dB. Compared to vanilla 3DGS, the student (Base) model—trained via multi-teacher distillation—preserves comparable reconstruction quality while significantly reducing the number of Gaussians. Although the student (Big) model achieves higher PSNR, it uses nearly as many Gaussians as the teacher models. In contrast, the student (Small) model applies further pruning, resulting in only a slight PSNR drop of 0.15 dB.

\noindent\textbf{Impact of different teachers.} We analyze the effects of various teacher models on the performance of the student model. As shown in Table~\ref{tab:abla_combine}, employing multiple diverse teachers ($G_{std}, G_{perb}, G_{drop}$) to distill the student yields the best overall performance. In contrast, using three standard teachers($G_{std}$) results in a lower PSNR (31.36), and single-teacher configurations perform even worse compared to these teacher ensembles. These results highlight that diversity among teachers provides richer and more complementary supervisory signals, thereby enhancing student model performance.
\section{Conclusion}
\label{sec:conclusion}

\noindent In this paper, we proposed a multi-teacher distillation framework for 3DGS, aiming to preserve reconstruction quality under significantly reduced Gaussian counts. By leveraging knowledge from multiple teacher models, our approach effectively transfers both scene geometry and appearance priors to a more compact student representation. Besides, we leverage a spatial distribution distillation strategy to encourage the student to learn spatial geometric distributions consistent with those of a standard teacher model. Extensive experiments across different scenes demonstrate that our distilled student model-Distilled-3DGS achieves promising performance with substantially fewer Gaussians, highlighting the potential of our method for deployment in memory-constrained or real-time scenes.

\noindent\textbf{Limitation.} Distilled-3DGS also has some drawbacks: first, it requires pre‑training multiple high‑performance teacher models, increasing training time and computational resources by at least N‑fold compared to a single model; second, generating distillation soft labels via multi‑model inference significantly increases GPU memory usage. Future work could explore end-to-end distillation pipelines or adaptive pruning strategies for Gaussian parameters to further improve efficiency and generalization.

\section*{Acknowledgement}
\label{sec:Acknowledgmen}
The computational resources are supported by SongShan Lake HPC Center (SSL-HPC) in Great Bay University. This work was also supported by Guangdong Research Team for Communication and Sensing Integrated with Intelligent Computing (Project No. 2024KCXTD047).

{
    \small
    \bibliographystyle{ieeenat_fullname}
    \bibliography{main}
}

\clearpage
\appendix



\begin{figure*}[t!]
\begin{center}
\includegraphics[width=1.0\textwidth]{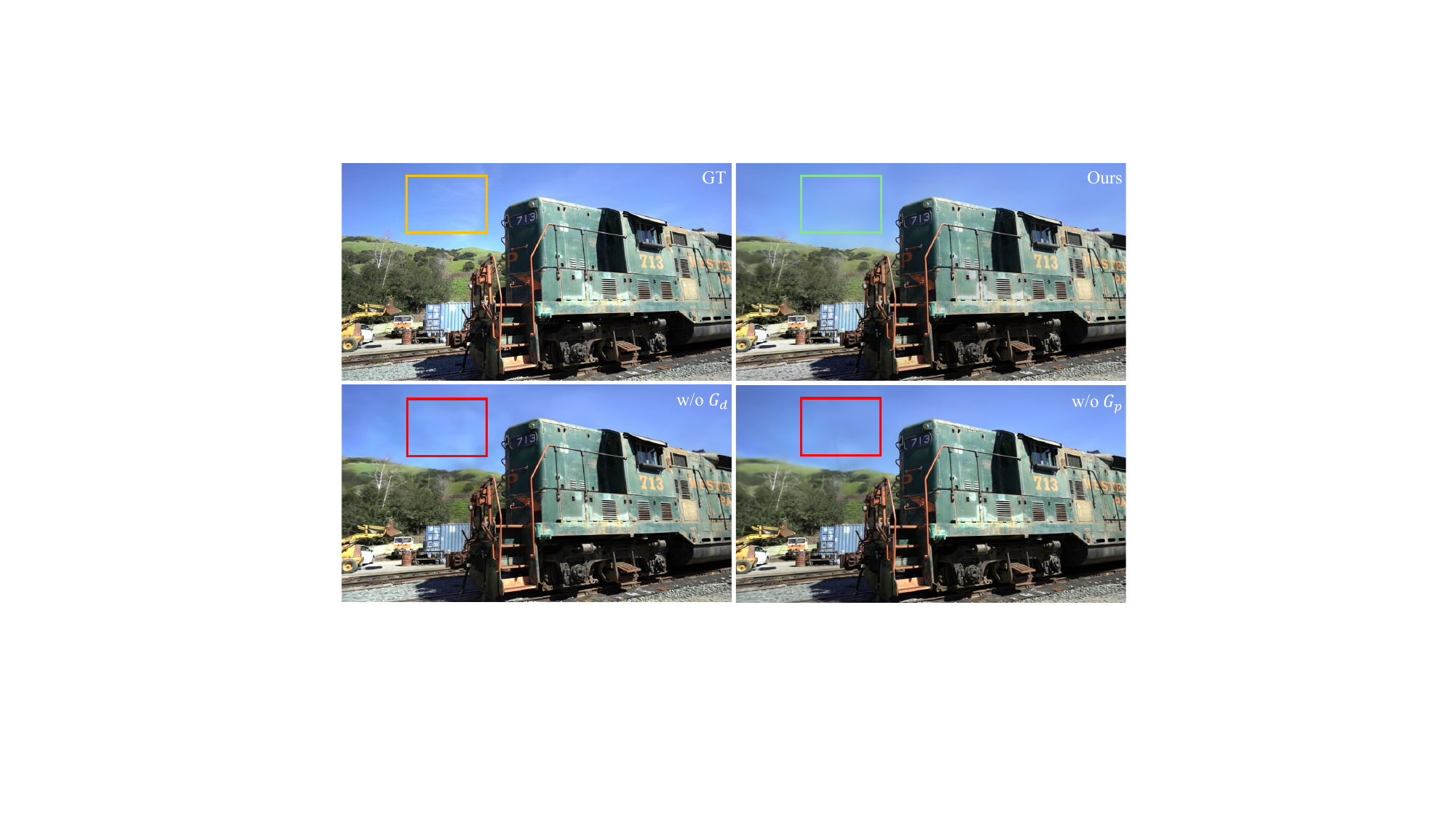}\\
    \caption{Visual comparison with different teacher models. Without the guidance of diverse teacher models, the rendering quality of the student 3DGS model gradually deteriorates.
  }
   \label{fig:abla_tea}
\end{center}
\end{figure*}

\section{Additional ablation experiments}
\subsection{Point cloud structural similarity}

Methods for evaluating the similarity of spatial geometric distribution between two point clouds can be broadly categorized into three types: distance-based, distribution matching-based, and learning-based approaches. Our proposed spatial distribution distillation strategy, which utilizes voxel histograms, falls under the distribution matching category. For the distance-based approach, we employ Chamfer Distance (CD) to directly measure the spatial distance between the two point sets. For the learning-based approach, we adopt Sonata~\cite{wu2025sonata}, a state-of-the-art point cloud representation learning method, to extract point features and compute the similarity loss between the two point clouds.

\begin{table}[htbp]
    \centering
    \resizebox{0.99\columnwidth}{!}{
    \begin{tabular}{l|ccccc}
    \toprule
    \multirow{2}{*}{\textbf{Method}} & \multicolumn{5}{c}{\textbf{Tanks\&Temples}} \\
    \cmidrule(lr){2-6} 
     & PSNR↑ & SSIM↑ & LPIPS↓  &Mem.(GB) &Student Time(min)  \\
 
    \midrule
    Distance-based &23.69 &0.832 &0.187 &23.82 & 50\\
    Feature-based &23.78 &0.847  &0.178  &40.00  &60\\
    \midrule
    \textbf{Ours} &23.76 &0.845 &0.179 &13.83  & 30\\
    \bottomrule
    \end{tabular}
    }
    \caption{Impact of different structural similarity strategy.}
    \label{tab:abla_dist}
\end{table}

As shown in Table~\ref{tab:abla_dist}, both the distance-based and feature-based methods consume considerable memory and time(student model training) yet fail to deliver substantial performance gains. In contrast, our proposed voxel histogram-based method outperforms these two approaches while requiring significantly less memory and computation time.

\section{Per-scene breakdown results}
To provide a more detailed evaluation of our model, we present the per-scene breakdown results of Mip-NeRF360, Tanks\&Temples and Deep Blending datasets. 
\begin{table}[t]
\centering
\resizebox{0.9\columnwidth}{!}{
\begin{tabular}{l l c c c c c c c}
\toprule
\multirow{1}{*}{\textbf{Scene}} & \multirow{1}{*}{\textbf{Method}} & \textbf{PSNR} & \textbf{SSIM} & \textbf{LPIPS}& \textbf{Num.(M)} \\
\midrule

\multirow{4}{*}{\textit{Bicycle}} 
 & EAGLES & \textbf{25.04} & 0.750 & \underline{0.240} & 2.26 \\
 &Mini-Splatting &\textbf{25.21}	&\underline{0.760} &0.246 &\textbf{0.60} \\
 & 3D-GS$^*$   &25.03 &0.740	&0.241 &5.67 \\
 &Ours &24.97 &\textbf{0.777} 	&\textbf{0.233} 	&\underline{0.59} \\
 
 \midrule
\multirow{4}{*}{\textit{Bonsai}} 
 & EAGLES      & 31.32 & 0.940 & 0.190  & 0.64 \\
 &Mini-Splatting &31.41	&0.940	&0.182	&\textbf{0.33} \\
 & 3D-GS$^*$ &\underline{31.99}	&\textbf{0.960}	&\textbf{0.170}	&1.64 \\
 &Ours &\textbf{32.79} 	&\underline{0.946} 	&\underline{0.179} 	&\underline{0.31} \\
 
 \midrule
\multirow{4}{*}{\textit{Counter}} 
 & EAGLES & 28.40 & 0.900 & 0.200  & 0.56 \\
 &Mini-Splatting &28.32	&\underline{0.913}	&\underline{0.181}	&\underline{0.36} \\
 & 3D-GS$^*$ &\underline{28.89}	&\textbf{0.920}	&0.190	&1.58 \\
 &Ours &\textbf{29.55} 	&0.914 	&\textbf{0.181} 	&\textbf{0.35} \\

 \midrule
\multirow{4}{*}{\textit{Flowers}} 
 & EAGLES & 21.29 & 0.58 & 0.370  & 1.33 \\
 &Mini-Splatting &\underline{21.31}	&\underline{0.614}	&\underline{0.334}	&\textbf{0.62} \\
 & 3D-GS$^*$ &21.30	&0.600	&0.359	&3.67  \\
 &Ours &\textbf{21.45} 	&\textbf{0.617} 	&\textbf{0.313} 	&\textbf{0.62} \\
 
 \midrule
\multirow{4}{*}{\textit{Garden}} 
 & EAGLES & 26.91 & 0.840 & 0.150 & 1.65 \\
 &Mini-Splatting &26.67	&0.844	&0.153	&\underline{0.69} \\
 & 3D-GS$^*$ &\underline{27.32}	&\underline{0.870}	&\underline{0.125}	&5.92 \\
 &Ours &\textbf{27.58} 	&\textbf{0.871} 	&\textbf{0.108} 	&\textbf{0.68} \\
 
 \midrule
\multirow{4}{*}{\textit{Kitchen}} 
 & EAGLES & 30.77 & 0.930 & 0.130 & 1.00 \\
 &Mini-Splatting &31.24	&0.924	&0.123	&\underline{0.38} \\
 & 3D-GS$^*$ &\underline{31.43}	&\underline{0.930} &\underline{0.120} &2.01 \\
 & Ours &\textbf{31.65} 	&\textbf{0.932} 	&\textbf{0.117} 	&\textbf{0.37} \\
 
\midrule
\multirow{4}{*}{\textit{Room}} 
 & EAGLES & 31.47 & 0.920 & 0.200 & 0.67 \\
 &Mini-Splatting &31.21	&0.920	&\underline{0.191}	&\underline{0.32} \\
 &3D-GS$^*$ &\underline{31.59}	&\underline{0.920}	&0.200	&1.99\\
 &Ours &\textbf{31.54} 	&\textbf{0.927} 	&\textbf{0.193} 	&\textbf{0.31} \\
 
 \midrule
\multirow{4}{*}{\textit{Stump}} 
 & EAGLES & 26.78 & 0.770 & 0.240  & 2.22 \\
 &Mini-Splatting &\underline{27.32}	&\underline{0.804}	&\underline{0.215}	&\underline{0.61} \\
 & 3D-GS$^*$ &26.53	&0.770	&0.240	&4.68\\
 &Ours &\textbf{27.73} 	&\textbf{0.811} 	&\textbf{0.193} 	&\textbf{0.60}  \\
 
 \midrule
\multirow{4}{*}{\textit{Treehill}} 
 & EAGLES & \underline{22.69} & 0.640 & 0.340 & 1.60 \\
 &Mini-Splatting &22.58	&\underline{0.656}	&0.331	&\underline{0.63} \\
 & 3D-GS$^*$ &22.43	&\textbf{0.660}	&\underline{0.325}	&3.67 \\
 &Ours &\textbf{22.98} 	&0.645 	&\textbf{0.314} 	&\textbf{0.62} \\ 

\midrule
\multirow{4}{*}{Average} 
 & EAGLES & 27.23 &0.810 & 0.240 & 1.33 \\
 &Mini-Splatting &27.25	&\underline{0.820} 	&\textbf{0.217} 	&\underline{0.50} \\ 
 & 3D-GS$^*$ &\underline{27.39}	&0.819	&\underline{0.219}	&3.43 \\
 &Ours &\textbf{27.81} 	&\textbf{0.827} 	&0.202 	&\textbf{0.49}  \\
 
\bottomrule
\end{tabular}
}
\caption{Quantitative per-scene breakdown results on MiP-NeRF360 dataset.}
\end{table}

\begin{figure}[H]
  \centering

\resizebox{1.0\columnwidth}{!}{
\begin{tabular}{l l c c c c c c c}
\toprule
\multirow{1}{*}{\textbf{Scene}} & \multirow{1}{*}{\textbf{Method}} & \textbf{PSNR$\uparrow$} & \textbf{SSIM}$\uparrow$ & \textbf{LPIPS}$\downarrow$& \textbf{\#G($10^6$)}$\uparrow$ \\

\midrule
\multirow{4}{*}{\textit{Playroom}} 
 & EAGLES &30.38 &0.910 &0.250 	&0.80 \\
 &Mini-Splatting &\textbf{30.62}  &\underline{0.915} &0.249 	&\underline{0.41} \\
 & 3D-GS$^*$   &28.79	&0.911 	&\textbf{0.241} 	&3.39 \\
 &Ours &\underline{30.45} 	&\textbf{0.926} 	&\underline{0.243} &\textbf{0.26} \\
 
\midrule
 \multirow{4}{*}{\textit{Johnson}} 
 & EAGLES &29.35 	&0.900 	&\textbf{0.240} 	&1.57 \\
 &Mini-Splatting &\underline{29.36} 	&0.903 	&0.260 	&\textbf{0.38}  \\
 & 3D-GS$^*$   &\textbf{30.31}	&\textbf{0.913} 	&\underline{0.241} 	&3.08 \\
 &Ours &29.29 	&\underline{0.906} 	&0.259 &\underline{0.39}  \\
\midrule
 \multirow{4}{*}{\textit{Average}} 
 & EAGLES &29.86 	&0.910 	&\underline{0.250} 	&1.19 \\
 &Mini-Splatting &\textbf{29.99} 	&0.909 	&0.255 	&\underline{0.40}   \\
 & 3D-GS$^*$   &29.55 	&\underline{0.912} 	&\textbf{0.241} 	&3.24  \\
 &Ours &\underline{29.87} 	&\textbf{0.916} 	&0.251 &\textbf{0.33}  \\
\bottomrule
\end{tabular}
}
\caption{Quantitative per-scene breakdown results on Deep Blending dataset.}
\end{figure}

\begin{table}[htbp]
\centering
\resizebox{0.99\columnwidth}{!}{
\begin{tabular}{l l c c c c c c c}
\toprule
\multirow{1}{*}{\textbf{Scene}} & \multirow{1}{*}{\textbf{Method}} & \textbf{PSNR}$\uparrow$ & \textbf{SSIM}$\uparrow$ & \textbf{LPIPS}$\downarrow$& \textbf{\#G($10^6$)}$\uparrow$ \\

\midrule
\multirow{4}{*}{\textit{Train}} 
 & EAGLES &21.65 &0.800 &0.240 	&0.46 \\
 &Mini-Splatting &21.28 &0.801 	&0.238 	&\underline{0.29}  \\
 & 3D-GS$^*$  &\underline{21.94}	&\underline{0.810} 	&\textbf{0.200} 	&1.11\\
 &Ours &\textbf{22.14} &\textbf{0.812} &\underline{0.207} &\textbf{0.23} \\
 
\midrule
 \multirow{4}{*}{\textit{Truck}} 
 & EAGLES &25.09 	&0.870 	&\underline{0.160} 	&0.83 \\
 &Mini-Splatting  &25.13 	&0.871 	&0.166 	&\underline{0.35}  \\
 & 3D-GS$^*$  &\underline{25.31}	&\textbf{0.880} 	&\textbf{0.150} 	&2.54 \\
 &Ours  &\textbf{25.37} 	&\underline{0.878} 	&\textbf{0.150} 	&\textbf{0.29}  \\

\midrule
 \multirow{4}{*}{\textit{Average}} 
 &EAGLES &23.37 &0.835 	&0.200 	&0.64  \\
 &Mini-Splatting &23.21 	&\underline{0.836} 	&0.203 	&\underline{0.32}  \\
 &3D-GS$^*$  &\underline{23.62} 	&\textbf{0.845} 	&\textbf{0.175} 	&1.83  \\
 &Ours &\textbf{23.76} 	&\textbf{0.845} 	&\underline{0.179} 	&\textbf{0.25}  \\
 
\bottomrule
\end{tabular}
}
\caption{Quantitative per-scene breakdown results on Tanks\&Temples dataset.}
\end{table}

\end{document}